\documentclass[letterpaper]{article} 
\usepackage{amsmath}
\usepackage{amssymb}
\usepackage{aaai25}  
\usepackage{times}  
\usepackage{helvet}  
\usepackage{courier}  
\usepackage[hyphens]{url}  
\usepackage{graphicx} 
\usepackage{duckuments}
\usepackage{booktabs}
\usepackage{subcaption}
\usepackage{natbib}  
\usepackage{caption} 
\usepackage{algorithm}
\usepackage{algorithmic}
\usepackage{color}

\usepackage{newfloat}
\usepackage{listings}
\DeclareCaptionStyle{ruled}{labelfont=normalfont,labelsep=colon,strut=off} 
\floatstyle{ruled}
\newfloat{listing}{tb}{lst}{}
\floatname{listing}{Listing}
\title{Open 3D World in Autonomous Driving}
\author{
    Xinlong Cheng\textsuperscript{\rm 1}\equalcontrib,
    Lei Li\thanks{Corresponding Author.}\textsuperscript{\rm 2}\equalcontrib
}
\affiliations{
    \textsuperscript{\rm 1}{Beijing University of Posts and Telecommunications} \\
     \textsuperscript{\rm 2}{University of Washington, University of Copenhagen} \\
    cxl.d5235252@gmail.com, lilei@di.ku.dk

}

\usepackage{bibentry}

\begin{document}

\maketitle

\begin{abstract}






The capability for open vocabulary perception represents a significant advancement in autonomous driving systems, facilitating the comprehension and interpretation of a wide array of textual inputs in real-time. Despite extensive research in open vocabulary tasks within 2D computer vision, the application of such methodologies to 3D environments, particularly within large-scale outdoor contexts, remains relatively underdeveloped. This paper presents a novel approach that integrates 3D point cloud data, acquired from LIDAR sensors, with textual information. The primary focus is on the utilization of textual data to directly localize and identify objects within the autonomous driving context. We introduce an efficient framework for the fusion of bird's-eye view (BEV) region features with textual features, thereby enabling the system to seamlessly adapt to novel textual inputs and enhancing the robustness of open vocabulary detection tasks. The effectiveness of the proposed methodology is rigorously evaluated through extensive experimentation on the newly introduced NuScenes-T dataset, with additional validation of its zero-shot performance on the Lyft Level 5 dataset. This research makes a substantive contribution to the advancement of autonomous driving technologies by leveraging multimodal data to enhance open vocabulary perception in 3D environments, thereby pushing the boundaries of what is achievable in autonomous navigation and perception.

\end{abstract}

%

\section{Introduction}

The field of autonomous driving has witnessed significant advancements, driven by the imperative for vehicles to accurately perceive and interpret their surroundings. Traditional perception systems have predominantly relied on visual data acquired from cameras \cite{liu2020smoke,chen2023ssflownet,wang2021fcos3d, chen2024cmu}. However, these systems face substantial limitations due to environmental factors such as lighting variations and adverse weather conditions. Recent innovations in open vocabulary tasks \cite{cheng2024yolo, chen2021groundingdino, li2024cpseg} have led to considerable progress in 2D computer vision, enhancing the ability of systems to comprehend and interpret a wide range of textual inputs. Despite these advancements, the application of open vocabulary perception within 3D environments \cite{boudjoghra20243d, lu2023open, vobecky2024pop}, especially in large-scale outdoor scenarios, remains a relatively unexplored and nascent area of research.

\begin{figure}[h]
   \centering
   \includegraphics[width=\linewidth]{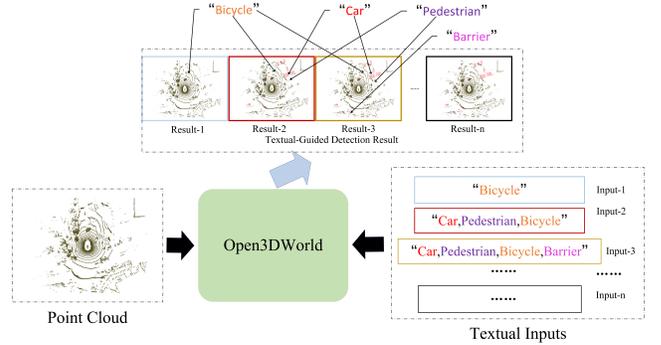}
   \caption{Unlike traditional closed-set 3D detection tasks, our Open3DWorld takes custom text as input and can locate and identify objects related to the text. Furthermore, new text inputs can be added seamlessly.}
   \label{fig:moti}
 \end{figure}



To address these limitations, the integration of LIDAR (Light Detection and Ranging) technology has become increasingly prevalent. LIDAR sensors generate detailed 3D point clouds, providing rich spatial information that enhances the perception capabilities of AVs \cite{zhou2018voxelnet, lang2019pointpillars, shi2022pillarnet, li2023pillarnext, yin2021center, li2023hierarchical}. Despite its advantages, point cloud data can be noisy and sparse \cite{yan2018second, fan2023fsd, fan2022fully, sun2022swformer, chen2023voxelnext}, presenting challenges for accurate object detection \cite{xu2022behind}. Moreover, relying solely on point cloud data may not fully exploit the available information in an autonomous driving scenario.

In addition to visual and spatial data, textual information plays a crucial role in understanding the driving environment \cite{kumar2024drive, xu2023drivegpt4, shao2024lmdrive}. Textual data from traffic signs, road markings, and external databases can provide contextual information that is essential for making informed driving decisions. However, the integration of text with spatial data for object detection in autonomous driving remains an underexplored area \cite{vobecky2024pop}.

On the other hand, current 3D point cloud methods \cite{chen2023voxelnext, sun2022swformer, fan2022fully,oehmcke2024deep}, especially in autonomous driving scenarios, have very limited detection categories due to the high cost of training data collection. Looking at image detection, open vocabulary detection methods such as YoloWorld \cite{cheng2024yolo, chen2021groundingdino} and GroundingDINO have shown good results and are expected to be applied in actual production and daily life. In autonomous driving scenarios, open vocabulary detection is of great significance because it can include numerous corner cases, ensure driving safety, and reduce costs.

%

This paper introduces a novel approach that integrates 3D point clouds from LIDAR sensors with textual data to enhance perception capabilities in autonomous driving. By directly using text to locate and identify objects, we propose a simple yet efficient method for fusing bird's-eye view (BEV) region features with text features. Our approach supports seamless adaptation to new textual inputs, facilitating robust open vocabulary detection tasks in 3D environments. We name our method \textquotedblleft{\textbf{Open3DWorld}}\textquotedblright.
The contributions of this paper are threefold:

\begin{enumerate}

    \item Our approach supports 3D open vocabulary detection tasks with LIDAR-text, allowing autonomous driving systems to seamlessly adapt to new textual inputs without requiring extensive retraining. This capability is essential for operating in diverse and dynamic environments.
    
    \item We propose a novel method for integrating 3D point clouds with textual data, enhancing the perception capabilities of autonomous driving systems. By efficiently fusing bird's-eye view (BEV) region features with text features, our method enables accurate object localization and identification directly from textual inputs.

    \item We have demonstrated the efficacy of our approach through comprehensive experiments on our expanded vocabulary of the NuScenes dataset \cite{caesar2020nuscenes}, termed as NuScenes-T dataset, and validated its zero-shot performance on the Lyft Level 5 dataset. 

\end{enumerate}

By advancing the integration of multimodal data for object detection, this work aims to enhance the safety, reliability, and efficiency of autonomous driving systems. 

\section{Related Work}

\paragraph{Open vocabulary}
Open vocabulary perception is an essential capability for autonomous driving systems, enabling the recognition and interpretation of a diverse range of textual inputs that may not be present in the training data. Recent advancements in this area have been facilitated by methods such as YoloWorld and GroundingDINO. YoloWorld is an advanced extension of the YOLO \cite{redmon2016you} framework, designed specifically to handle open vocabulary tasks by integrating contextual information from large-scale language models \cite{cheng2024yolo}. GroundingDINO combines the grounding of textual descriptions with a visual perception model, enabling precise alignment of text and image data \cite{chen2021groundingdino}.
To train and validate models for open vocabulary perception, several large-scale and diverse datasets have been utilized, including COCO (Common Objects in Context) \cite{lin2014coco}, Objects365 (O365) \cite{shao2019objects365}, Golden Gate Dataset (GoldG), and Conceptual Captions 3 Million (CC3M) \cite{sharma2018conceptual}. These datasets provide a wide variety of object categories and annotations, facilitating comprehensive training for 2D open vocabulary tasks.
For 3D open vocabulary task,3D-OWIS\cite{boudjoghra20243d} proposes a new open-world 3D indoor instance segmentation method by automatically annotating and generating pseudo labels and adjusting unknown category probabilities to distinguish and gradually learn unknown categories.OV-3DET\cite{lu2023open} proposes a method that complete open vocabulary point cloud object detection without any 3D annotations.In outdoor scene,POP-3D\cite{vobecky2024pop} predict 3D occupancy by using pretrained multimodal model.

\paragraph{3D Object Detection in Autonomous Driving}
VoxelNet \cite{zhou2018voxelnet} is the first to introduce dense convolutions for LiDAR-based 3D object detection, achieving competitive performance. PointPillars \cite{lang2019pointpillars}, PillarNet \cite{shi2022pillarnet}, and PillarNext \cite{li2023pillarnext} utilize 2D dense convolutions on BEV (bird's-eye view) dense feature maps. SECOND \cite{yan2018second}, a pioneering effort, employs a sparse CNN to extract 3D sparse voxel features and then transforms them into dense BEV feature maps for predictions. CenterPoint \cite{yin2021center} introduces a center-based detection head. FSDv1 \cite{fan2022fully} segments raw point clouds into foreground and background, then clusters the foreground points to represent individual objects. It uses a PointNet-style network to extract features from each cluster for initial coarse predictions, refined by a group correction head. FSDv2 \cite{fan2023fsd} replaces instance clustering with a virtual voxelization module, aiming to eliminate the inductive bias of handcrafted instance-level representations. SWFormer \cite{sun2022swformer} presents a fully transformer-based architecture for 3D object detection. More recently, VoxelNeXt \cite{chen2023voxelnext} streamlines the fully sparse architecture with a purely voxel-based design, localizing objects based on the features nearest to their centers.

\paragraph{Multimodality Fusion}For text and image feature alignment, CLIP \cite{radford2021learning} is a pioneering work in unsupervised cross-modal training. Subsequently, MaskCLIP \cite{dong2023maskclip} refines CLIP for pixel-level dense prediction tasks, particularly semantic segmentation.
Multi-modal feature fusion methods for open vocabulary detection include YoloWorld \cite{cheng2024yolo}, which proposes a Vision-Language PAN to fuse vocabulary embeddings and multi-scale image features. GroundingDINO \cite{chen2021groundingdino} introduces a Feature Enhancer and a Language-guided Query Selection module, employing an architecture similar to transformers.
For image and point cloud feature alignment, BEVFusion \cite{liang2022bevfusion} first converts image and point cloud data to the bird's-eye view (BEV) space and then fuses them.
For text, image, and point cloud alignment in indoor scenes, OV-3DET \cite{lu2023open} proposes a Debiased Cross-modal Triplet Contrastive loss. POP-3D \cite{vobecky2024pop} addresses outdoor scenes by using the pretrained MaskCLIP model and employing image features as a medium to establish connections between the three modalities.


\section{Problem Formulation}
In this section, we formally define the problem of 3D open vocabulary perception for autonomous driving. Our task involves integrating 3D point clouds from LIDAR sensors with textual data to enhance object localization and identification in large outdoor environments.

Let $\mathcal{P} = \{p_1, p_2, \ldots, p_n\}$ represent a set of 3D points obtained from LIDAR sensors, where each point $p_i \in \mathbb{R}^3$ is a coordinate in the 3D space. Additionally, let $\mathcal{T} = \{t_1, t_2, \ldots, t_m\}$ represent a set of textual inputs, where each text $t_j$ is a word or phrase describing an object or region of interest in the driving environment.

Our objective is to develop a function $F: (\mathcal{P}, \mathcal{T}) \rightarrow \mathcal{L}$ that maps the set of 3D points $\mathcal{P}$ and textual inputs $\mathcal{T}$ to a set of localized regions $\mathcal{L}$ in the 3D space. Each localized region $l_k \in \mathcal{L}$ is defined by a bounding box or a set of coordinates that encapsulate the identified object or region corresponding to the textual description.

Formally, the problem can be described by the following steps:
\begin{enumerate}
    \item \textbf{Feature Extraction:} Extract bird's-eye view (BEV) features $\mathcal{B}$ from the 3D point clouds $\mathcal{P}$. This can be represented as $\mathcal{B} = f_{\text{BEV}}(\mathcal{P})$, where $f_{\text{BEV}}$ is the feature extraction function.
    \item \textbf{Textual Feature Encoding:} Encode the textual inputs $\mathcal{T}$ into feature vectors $\mathcal{E} = \{e_1, e_2, \ldots, e_m\}$ using a text encoder $f_{\text{TE}}$. This can be represented as $e_j = f_{\text{TE}}(t_j)$ for each text $t_j \in \mathcal{T}$.
    \item \textbf{Feature Fusion:} Fuse the bird's-eye view (BEV) features $\mathcal{B}$ with the textual feature vectors $\mathcal{E}$ to generate a set of fused features $\mathcal{F}$. This can be represented as $\mathcal{F} = f_{\text{fusion}}(\mathcal{B}, \mathcal{E})$, where $f_{\text{fusion}}$ is the fusion function.
    \item \textbf{Localization and Identification:} Apply a localization function $f_{\text{loc}}$ on the fused features $\mathcal{F}$ to generate the set of localized regions $\mathcal{L}$. This can be represented as $\mathcal{L} = f_{\text{loc}}(\mathcal{F})$.
\end{enumerate}

Our goal is to design the functions $f_{\text{BEV}}$, $f_{\text{TE}}$, $f_{\text{fusion}}$, and $f_{\text{loc}}$ such that the overall function $F$ accurately maps the 3D point clouds and textual inputs to precise and meaningful localized regions in the 3D space, supporting robust open vocabulary perception in autonomous driving scenarios.

\section{Methodology}
\subsection{NuScenes-T Dataset}
\begin{figure}[H]
   \centering
   \includegraphics[width=\linewidth]{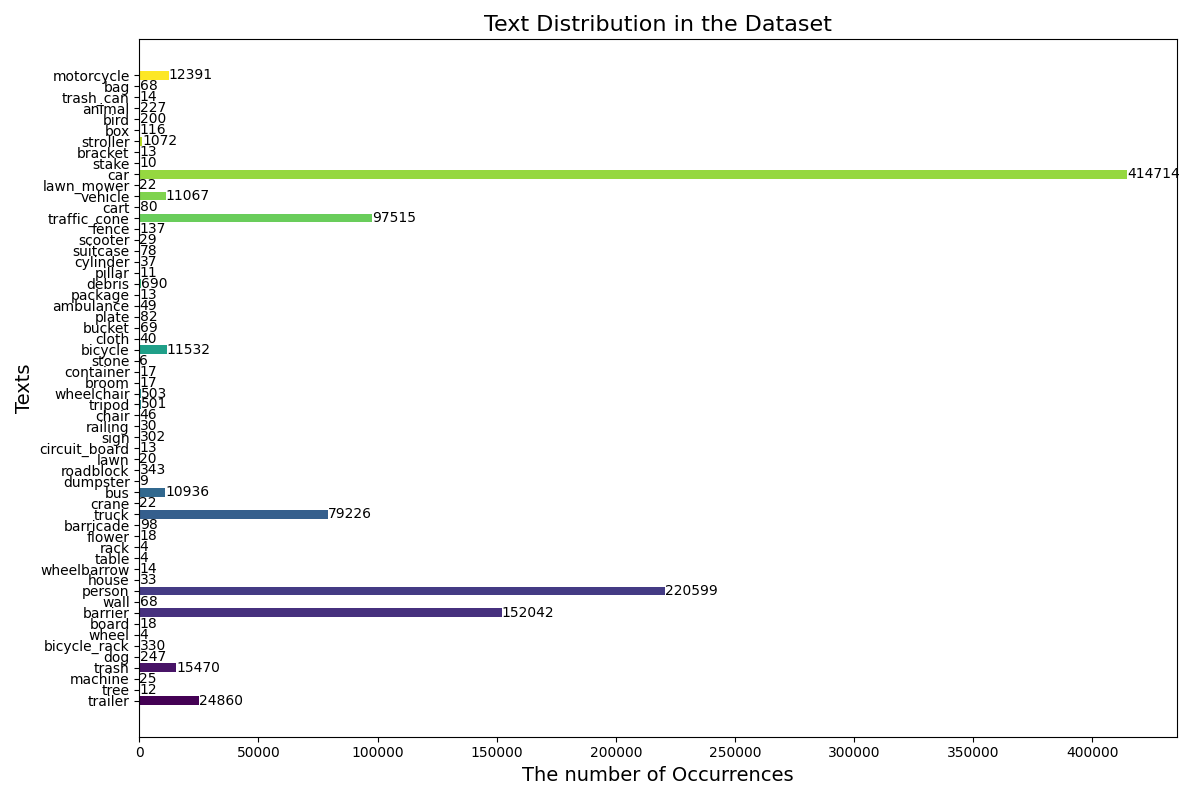}
   \caption{The distribution of different texts in the NuScenes-T dataset, showing the number of objects described by each noun.}
   \label{fig:data}
 \end{figure}

    TOD3Cap\cite{jin2024tod3cap} gives a detailed description of each object in NuScenes dataset based on the original annotations. We use it to extract the noun subjects of the descriptions, and after filtering, we get the noun descriptions of the objects in NuScenes, for example,\textquotedblleft{car}\textquotedblright,\textquotedblleft{box}\textquotedblright,\textquotedblleft{trash}\textquotedblright.The distribution of all nouns is shown in the Fig.\ref{fig:data}.


Through this method, we obtain many more category annotations than those in the original NuScenes dataset. Some common challenging cases in autonomous driving scenarios, such as \textquotedblleft{stone}\textquotedblright and \textquotedblleft{box}\textquotedblright, are also included in our new categories.

\begin{figure*}[th]
   \centering
   \includegraphics[width=\linewidth]{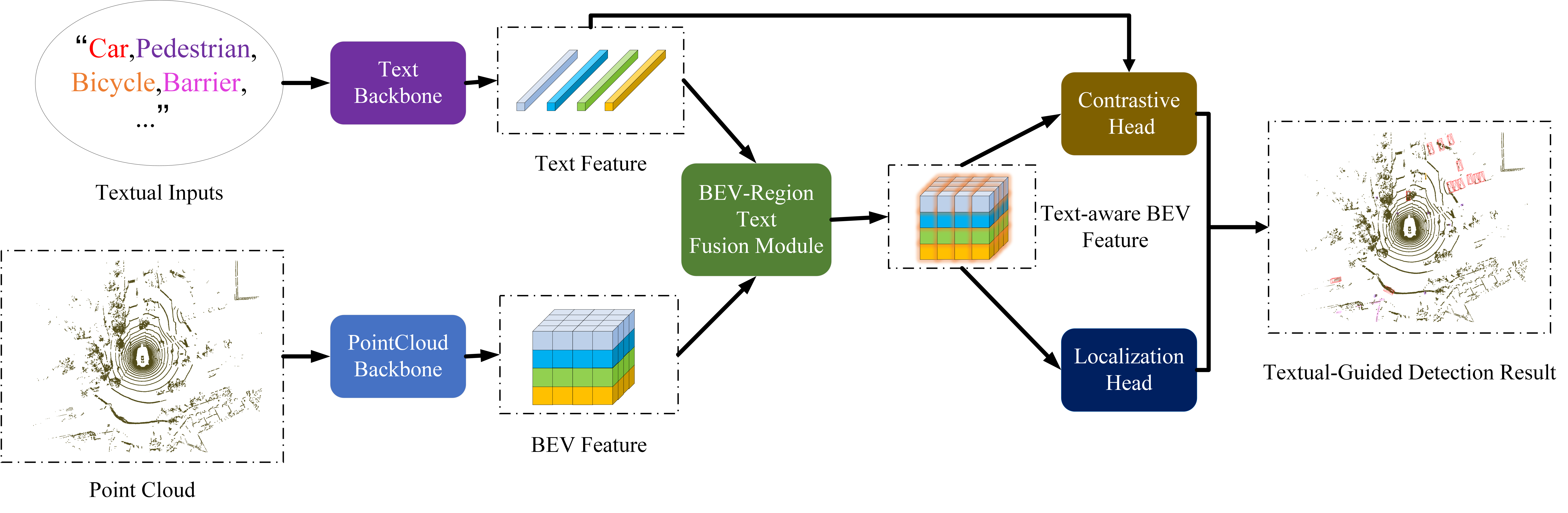}
   \caption{Framework Overview: We start by separately extracting text features and bird's-eye view (BEV) features. Then, we use the BEV-Region Text Fusion module to align these features, connecting the BEV features with the text features. This alignment allows us to obtain unified BEV and text features. Finally, a contrastive head calculates the similarity between BEV region features and text features to identify the relevant positions in 3D space based on the text input. A localization head then refines the object's location and recognition details.}
   \label{fig:over}
 \end{figure*}

\begin{figure}[t]
   \centering
   \includegraphics[width=0.7\linewidth]
   {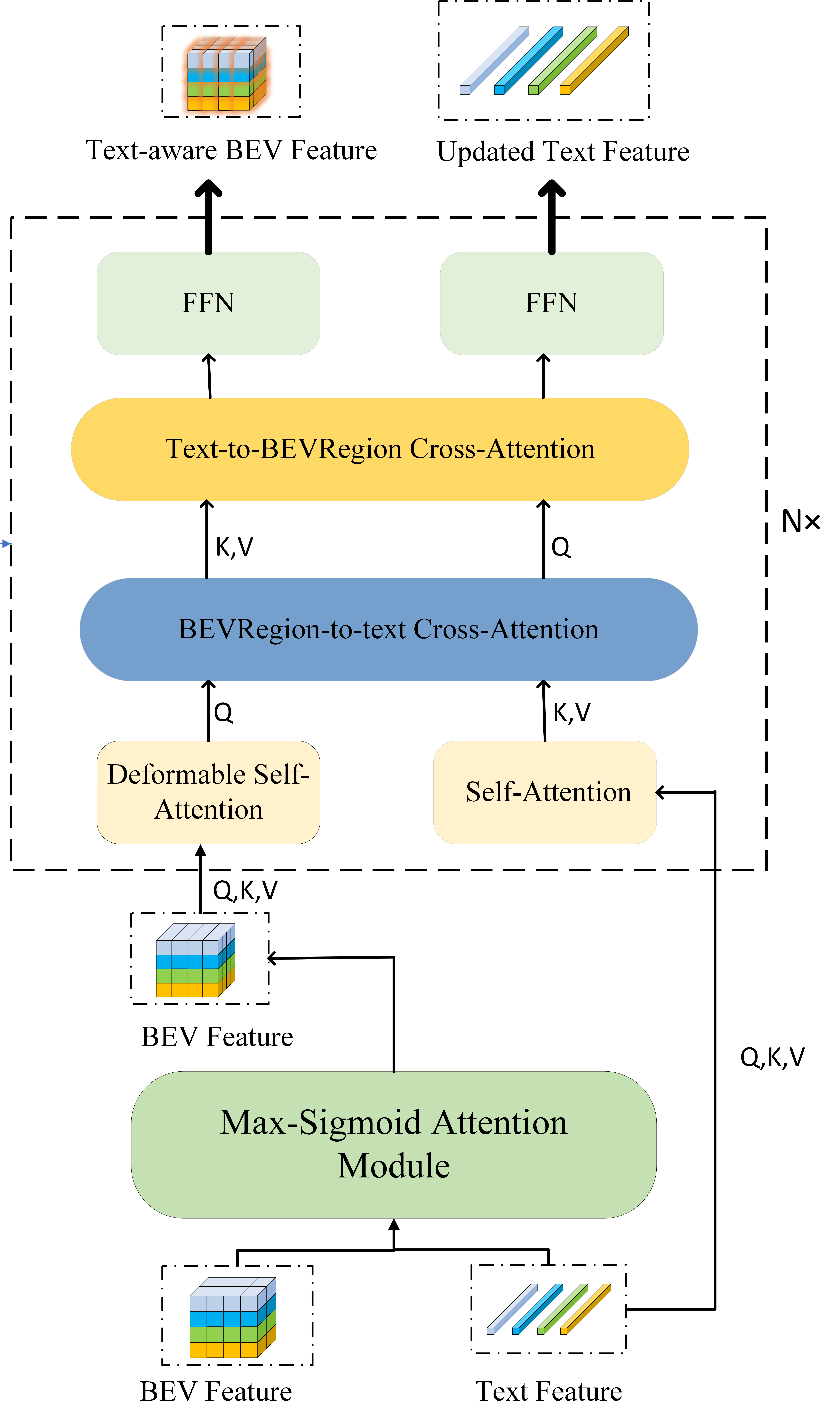}
   \caption{We propose a progressive alignment module to fuse vanilla BEV features and vanilla text features. It takes the features of the two modalities, extracted separately by the backbone, as inputs and outputs \textbf{Text-aware BEV feature} and \textbf{Updated text Feature}, which provide a crucial basis for determining the direct connections between BEV grids and texts. Our fusion module fully leverages the attention mechanism's capability to integrate data from different modalities.}
   \label{fig:fusion}
 \end{figure}

\begin{figure*}[htbp]  
    \centering  
    \begin{subfigure}[b]{0.49\linewidth}  
        \centering  
        \includegraphics[width=1.0\linewidth, height=0.4\linewidth]{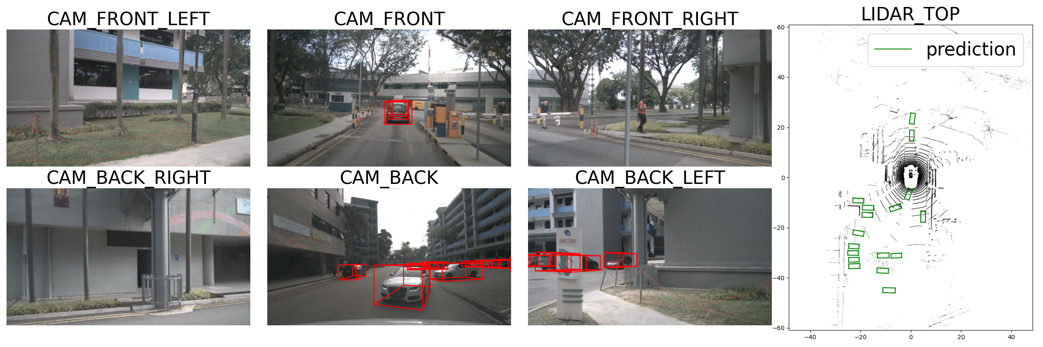}  
        \textbf{textual input:\textquotedblleft{car}\textquotedblright}
        \label{fig:sub-first}  
    \end{subfigure}  
    \hfill 
    \begin{subfigure}[b]{0.49\linewidth}  
        \centering  
        \includegraphics[width=1.0\linewidth, height=0.4\linewidth]{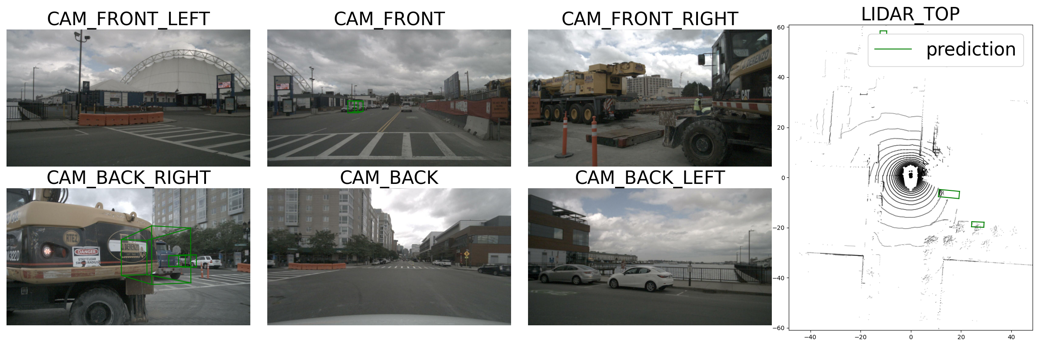}  
        \textbf{textual input:\textquotedblleft{truck}\textquotedblright}
        \label{fig:sub-second}  
    \end{subfigure}  
  
    \begin{subfigure}[b]{0.49\linewidth}  
        \centering  
        \includegraphics[width=1.0\linewidth, height=0.4\linewidth]{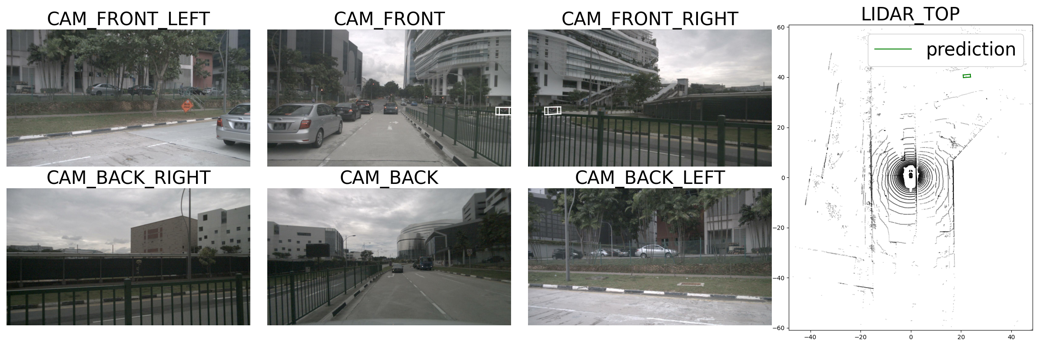}  
        \textbf{textual input:\textquotedblleft{motocycle}\textquotedblright}
        \label{fig:sub-third}  
    \end{subfigure}  
    \hfill  
    \begin{subfigure}[b]{0.49\linewidth}  
        \centering  
        \includegraphics[width=1.0\linewidth, height=0.4\linewidth]{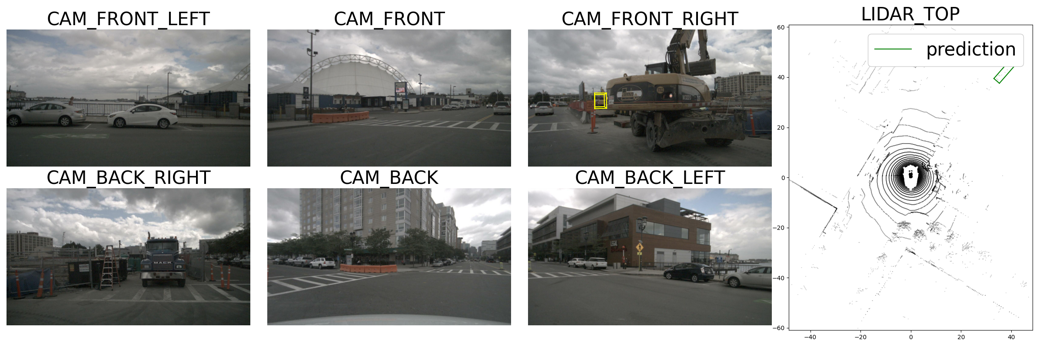}  
        \textbf{textual input:\textquotedblleft{bus}\textquotedblright}
        \label{fig:sub-fourth}  
    \end{subfigure}  
  
    \caption{NuScenes-T dataset’s results.The caption is the textual input. Based on the chosen textual data, Open3DWorld can output 3D boxes related to the text. Due to space limitations, additional visualization results are placed in the appendix. Note that our model takes only point cloud and text as input, with images used solely to display the detection results.}  
    \label{fig:nus}  
\end{figure*}  

\subsection{Framework Overview}
The overall framework of our proposed method is illustrated in Fig.\ref{fig:over} and mainly consists of three components. The first is the feature extraction backbone, which includes both text and point cloud feature extractors. The second component is the bird's-eye view (BEV) features and text features fusion module, which extracts text-aware BEV features and obtains updated text features. Finally, the last component is the multi-modal head, which consists of a contrastive head and a localization head. In summary, our method outputs the target 3D information related to the textual inputs, including the target’s spatial position, size, heading, and other details.

\subsection{Feature Extraction}
Our approach can be seamlessly adapted to different text and point cloud feature extractors.In our configuration,we adpot CLIP(Contrastive Language-Image Pre-Training)  text encoder as text backbone.For point cloud extractor,we propose a new module to obtain bird's-eye view (BEV) features based on SECOND, termed as \textquotedblleft{OpenSECOND}\textquotedblright.

We use all the words extracted from external information. We use a text encoder to get the embeddings of the textual inputs, and the textual inputs' embeddings are represented as $\mathcal{E} \in \mathbb{R}^{m \times d}$, which is used to prompt the content that needs to be localized and identified.

For point cloud, we use a point cloud encoder to get BEV features, as a feature space that interacts with the textual input's feature space. We first partition the 3D space into voxels, transforming raw point clouds into binary voxel grids. Subsequently, a 3D sparse convolutional neural network is applied to the voxel grid for feature representation. Similar to the image features mentioned earlier, Z-axis pooling yields the point cloud BEV featmap $\mathcal{B} \in \mathbb{R}^{H \times W \times d}$.

\subsection{BEV-Region Text Fusion Module}
When fusing multi-modal features, the fusion module is very important.In our Open3DWorld, We named it \textbf{BEV-Region Text Fusion Module} because it establishes the relationship between each BEV grid and each text and updates both to align their feature spaces.We conduct extensive experiments to explore the fusion module that best suits our task, and finally design the fusion module shown in the Fig.\ref{fig:fusion}.

First, we flatten the bird's-eye view (BEV) features to get a flattened features  $\mathcal{B}_f \in \mathbb{R}^{n\times d}$, and text features is $\mathcal{T}\in \mathbb{R}^{m\times d}$.In order to initially fuse the multi-modal features,we use Max-Sigmoid Attention Module to update BEV-region features by using text features,whic is formated as:

\begin{equation}  
\mathcal{B}_f \cdot \delta\left(\max_{t=1}^{m}(\mathcal{B}_f\mathcal{T}_t^{\top})\right)^{\top}  
\end{equation}

The  $\delta$  indicates the sigmoid function.After this, bird's-eye view (BEV) features initially have the ability of \textquotedblleft{text-aware}\textquotedblright.Next,we fuse BEV and text features more fully. We first perform self-attention on BEV features and text features respectively. Considering the computational burden brought by the large number of bird's-eye view (BEV) grid, we use deformable self-attention to reduce the computational burden. Next, we first use cross-attention to aggregate text features into bird's-eye view (BEV) features, and then use cross-attention to aggregate BEV features into text features. Finally, we use FFN to adjust the feature dimension. Like Encoder blocks in Transformer, this fusion process is performed N times. We have proved through experiments that N equals 3 to achieve a balance between effect and computational burden.

\subsection{Contrastive Head and Localization Head}
For Contrastive head, we use several convs to get final BEV grid's features.Next,We calculate the similarity between each BEV grid and text using the following formula,and $s_{i,j}$ represent the similarity between i-th BEV grid and j-th text:
\begin{equation}  
s_{i,j} = \alpha \cdot BN(b_{i} )\cdot BN(t_{j} )+\beta 
\end{equation}

where $BN(\cdot)$ is batch normalization.Following previous work\cite{cheng2024yolo}, we add the affine transformation with the learnable scaling factor $\alpha$ and shifting factor $\beta$.Both batch normalization and the affine transformations are important for stabilizing the BEV-region and text fusion training.

For Localization head,we are consistent with the traditional 3D detection tasks and regress all the information,including the target’s spatial position, size,heading and other information.Our advantage is that we predict information based on BEV features that are integrated with text features, so it is easier to predict attributes of a certain category.

\subsection{Training and Evaluation}
After obtaining the similarity map $\mathcal{S} \in \mathbb{R}^{n\times m}$ between BEV grid and text,we obtain GT heatmap $\mathcal{H} \in \mathbb{R}^{n\times m}$ in two steps.
Firstly, we project 3D GT boxes to the BEV feature map, which yields rotated boxes. For example, we project a 3D box \( B = [x, y, z, x_{\text{size}}, y_{\text{size}}, z_{\text{size}}, \varphi] \) into the BEV featmap:

\begin{equation}  
x_{bev}=\frac{x-R_{pc}}{F_{v}\times F_{o}},y_{bev}=\frac{y-R_{pc}}{F_{v}\times F_{o}}
\end{equation}
\begin{equation}
    x_{bev-size}=\frac{x_{size}}{F_{v}\times F_{o}},y_{bev-size}=\frac{y_{size}}{F_{v}\times F_{o}},
\end{equation}
where $F_{v},F_{o}$ and $R_{pc}$ represent the voxel factor,out size factor of the backbone.Secondly,we use a sample allocation method similar to YoloWorld to obtain $\mathcal{H}$.Finally,
we use cross entropy loss to supervise training,which is formated as:

\begin{equation}  
\mathcal{L}_{contrastive} = CrossEntropyLoss(\mathcal{S},\mathcal{H})
\end{equation}  

Localization loss $\mathcal{L}_{localization}$ is similar to CenterPoint, which use L1 loss for 3D bounding box regression. To balance two different losses, we set a weight coefficient $\alpha=0.025$. Finally,the total loss is

\begin{equation}  
\mathcal{L}_{total} = \mathcal{L}_{contrastive}+\alpha*\mathcal{L}_{localization}.
\end{equation} 

During the evaluation process, after generating the similarity map between BEV grids and the textual input, denoted as $\mathcal{S} \in \mathbb{R}^{n\times m}$, we apply a predefined score threshold. If the similarity score between a BEV grid and the corresponding text exceeds this threshold, the grid is deemed the center of the object described by the text. The object is then identified and localized by extracting the 3D information of the corresponding grid. Finally, we apply Non-Maximum Suppression (NMS) to filter and refine the detections.

\section{Experiments}


\subsection{Dataset}


We conducted our experiments on the NuScenes-T dataset and the Lyft Level 5 dataset \cite{houston2021one}. The NuScenes-T dataset is an extension of the NuScenes dataset, a large-scale benchmark designed for autonomous driving research. The NuScenes dataset comprises data collected from 1,000 real-world driving scenes, each approximately 20 seconds in duration. The dataset is partitioned into 700 scenes for training, 150 for validation, and 150 for testing. Notably, two keyframes are annotated per second across the entire dataset, resulting in a total of 1.4 million 3D bounding boxes. The Lyft Level 5 dataset, on the other hand, was introduced as part of the NeurIPS 2019 competition and serves as another critical resource for autonomous driving research.

\subsection{Experimental Setup}

We conducted experiments on the NuScenes-T and Lyft Level 5 datasets. We tested 10 key categories in NuScenes-T and evaluated zero-shot performance on Lyft Level 5 without any retraining. For training, we used all NuScenes-T samples and expanded categories with additional nouns from other datasets. We compared our model's performance on the NuScenes-T test set and then assessed zero-shot performance on the Lyft Level 5 test set. The point cloud range for both datasets was set to \([-54\, \text{m}, -54\, \text{m}, 54\, \text{m}, 54\, \text{m}]\). We used the CLIP text encoder for text embeddings and a redesigned point cloud backbone to extract BEV features, with feature dimensions set to 512. We applied cross-entropy loss for contrastive learning and L1 loss for localization. The model was trained with the AdamW optimizer and a CosineAnnealingLR scheduler over 20 epochs.

\subsection{Performance Metrics}
\begin{table*}[t]
\centering
\begin{tabular}{ll|ccccccccc}
    \toprule
    T.B. & P.B. & Params & FPS & mAP(↑) & NDS(↑)& mATE(↓) &mASE(↓)&mAOE(↓)&mAVE(↓)\\
    \midrule
    CLIP Text Encoder & PointPillars &  52M & 23&0.3975&0.5528&0.3342&0.2671&0.2911&0.2735\\
    CLIP Text Encoder & SSN &  130M&8&0.4413&0.5717&0.2916&0.2008&0.2529&0.2457\\
    CLIP Text Encoder & Centerformer &  177M&5&0.4622&0.6149&0.2872&0.2289&0.2437&0.2418\\
    BERT-base & OpenSECOND & 122M&9&0.4171&0.5862&0.3118&0.2589&0.2795&0.2519\\
    CLIP Text Encoder & OpenSECOND&76M&14&0.4455&0.5924&0.2920&0.2573&0.2881&0.2675\\
    
    \bottomrule
\end{tabular}
\caption{Results of ablation experiments.\textquotedblleft{T.B.}\textquotedblright means Text Backbone,\textquotedblleft{P.B.}\textquotedblright means Point cloud Backbone.}
\label{table:nus}
\end{table*}


\begin{table}[t]  
\centering  
\begin{tabular}{l|cccccc} 
    \toprule 
    Model & car & bus & truck & bi. & pe. & motor. \\  
    \midrule 
    P.P. & 0.422 & 0.177 & 0.176 & 0.099 & 0.066 & 0.049 \\  
    Ours & 0.279& 0.106 & 0.089 & 0.047 & 0.043 & 0.018 \\  
    \bottomrule 
\end{tabular}  
\caption{Ours Zero-shot vs. PointPillars(\textbf{trained on Lyft Level 5 dataset},abbreviated as P.P.).In the figure, \textquotedblleft{bi.}\textquotedblright represents \textquotedblleft{bicycle}\textquotedblright,\textquotedblleft{pe.}\textquotedblright represents \textquotedblleft{pedestrian}\textquotedblright and \textquotedblleft{motor.}\textquotedblright  represents \textquotedblleft{motorcycle}\textquotedblright. Lyft uses a method similar to COCO to compute the mean average precision (mAP) – calculating the average precision under different thresholds of 3D IoU ranging from 0.5 to 0.95. A 3D IoU overlap of more than 0.7 is a strict criterion, which can make overall performance appear lower.}  
\label{table:lyft}  
\end{table} 

In NuScenes-T dataset,we report the officially used metrics of 3D object detection in BEV-based research\cite{caesar2020nuscenes,lang2019pointpillars,li2023bevdepth}, i.e., mean Average Precision(mAP) and five True Positive metrics, including mean Average Translation Error (mATE), mean Average Scale Error (mASE), mean Average Orientation Error(mAOE), mean Average Velocity Error(mAVE), mean Average Attribute Error(mAAE), where the lower value is better. Besides, the NuScenes Detection Score (NDS) comprehensively reflects these metrics, and it is the most concerned metric for performance evaluation.

In Lyft Level 5 dataset, we adopts a way similar to COCO to compute the mean average precision (mAP) – compute the average precision under different thresholds of 3D IoU from 0.5-0.95.

\subsection{Results and Analysis}
With the vocabulary expanded by external description information, we train Open3DWorld on the NuScenes-T dataset, which allows it to complete detection tasks similarly to previous 3D detection models. Since our vocabulary includes the categories in the closed set for detection tasks, we can calculate quantitative indicators like those in standard 3D detection tasks.


Our detection accuracy is comparable to that of specialized 3D detection models, despite not employing additional training techniques to enhance performance. This is intentional, as our primary objective is to develop a general open vocabulary model capable of seamlessly integrating new textual information. The test results on the Nuscenes-T dataset are presented in Table.~\ref{table:nus}.

\begin{figure*}[h]
    \centering
    \begin{minipage}[t]{.49\linewidth}
        \centering
        \includegraphics[width=\linewidth, height=0.5\linewidth]{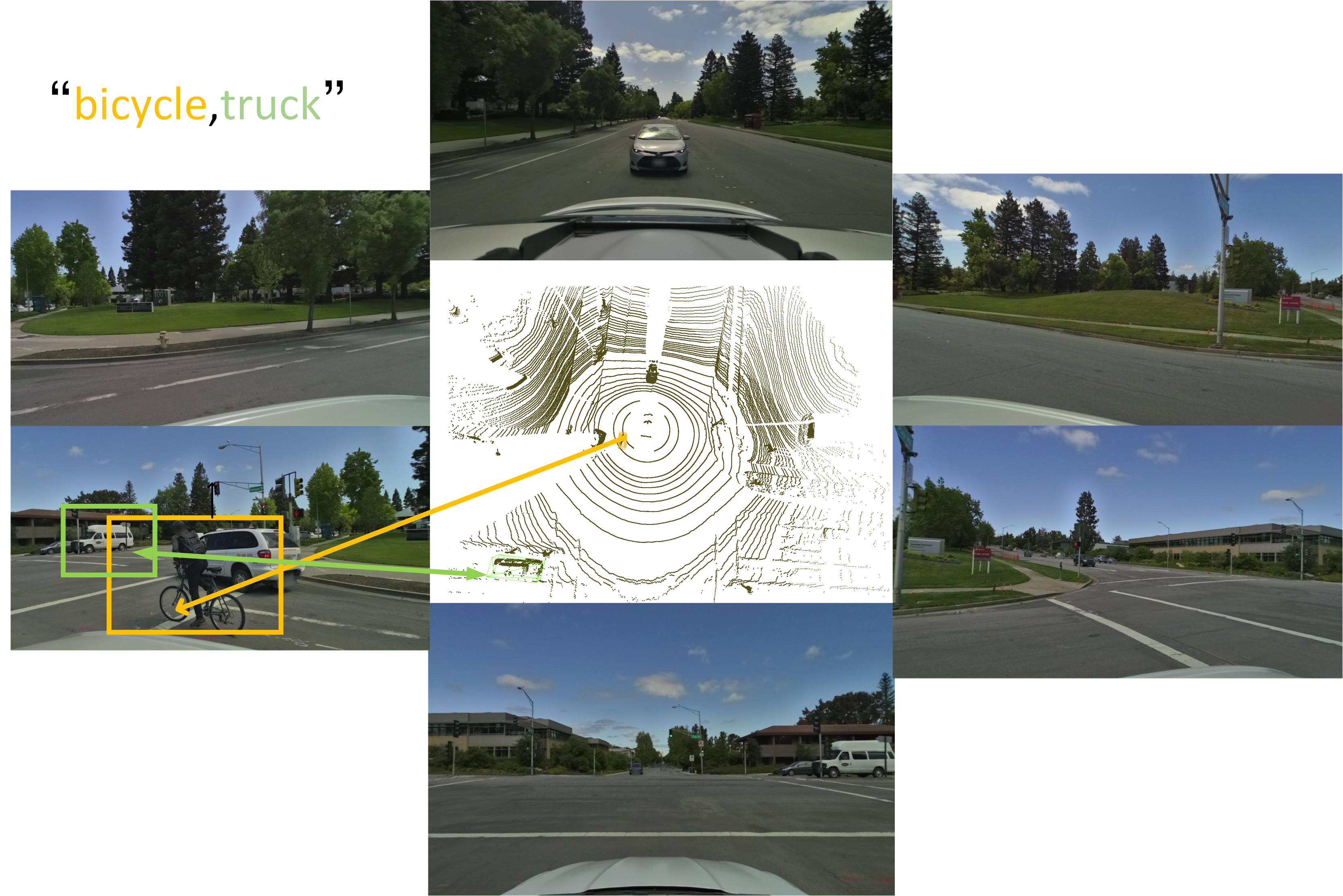}
        \label{fig:bi_tr}
    \end{minipage}\hfill
    \begin{minipage}[t]{.49\linewidth}
        \centering
        \includegraphics[width=\linewidth, height=0.5\linewidth]{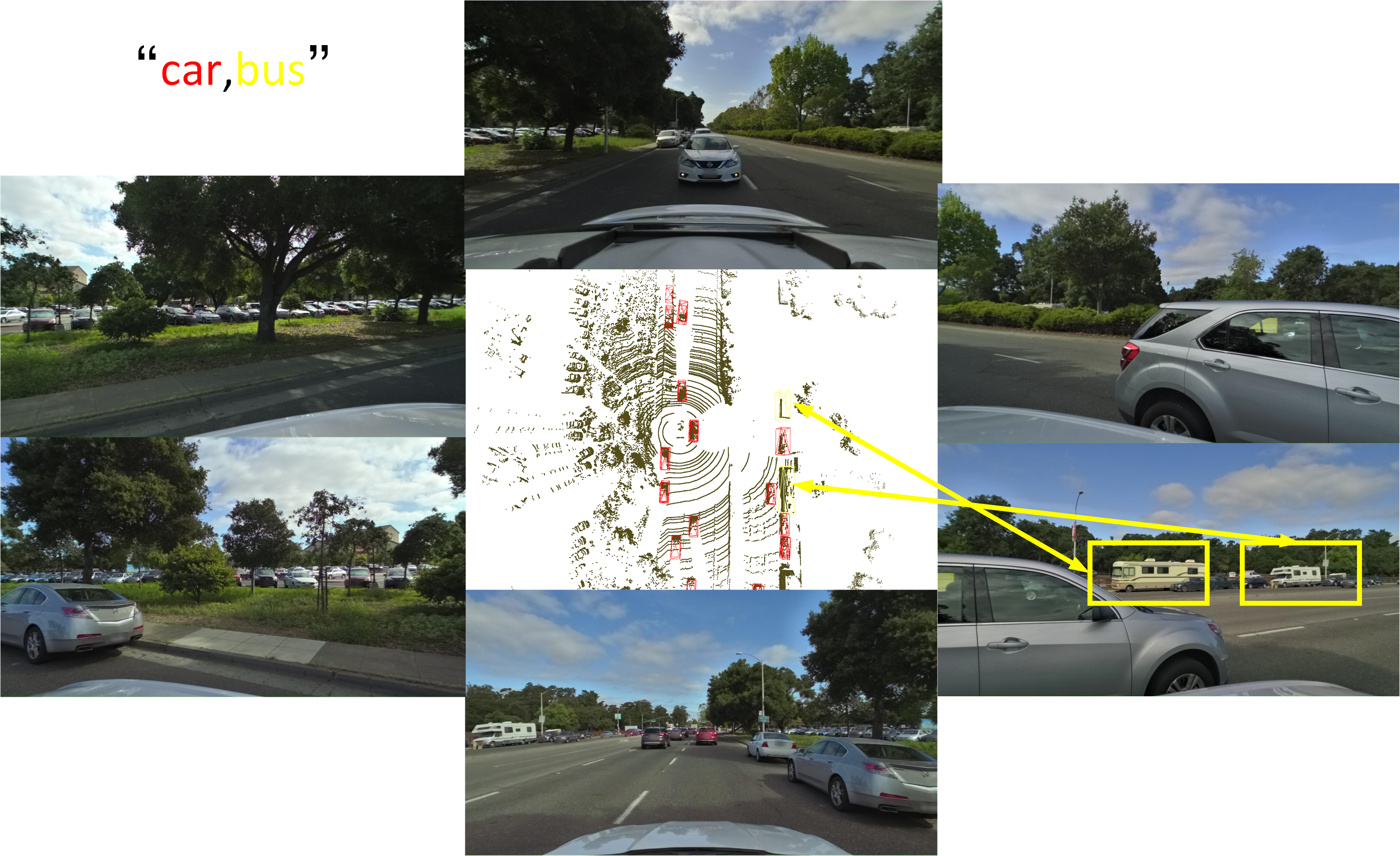}
        \label{fig:car_bus}
    \end{minipage}


    \begin{minipage}[t]{.49\linewidth}
        \centering
        \includegraphics[width=\linewidth, height=0.5\linewidth]{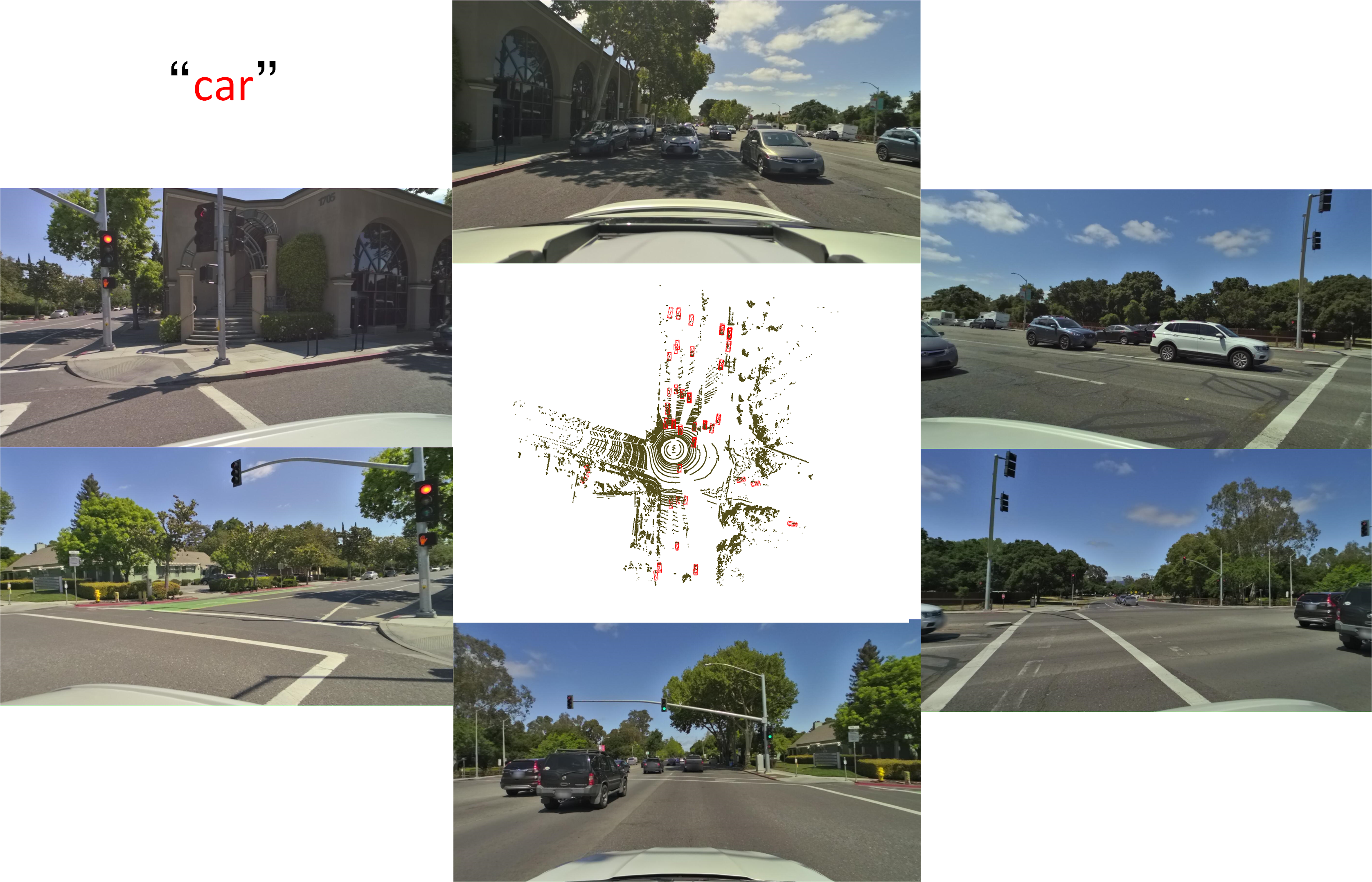}
        \label{fig:car}
    \end{minipage}\hfill
    \begin{minipage}[t]{.49\linewidth}
        \centering
        \includegraphics[width=\linewidth, height=0.5\linewidth]{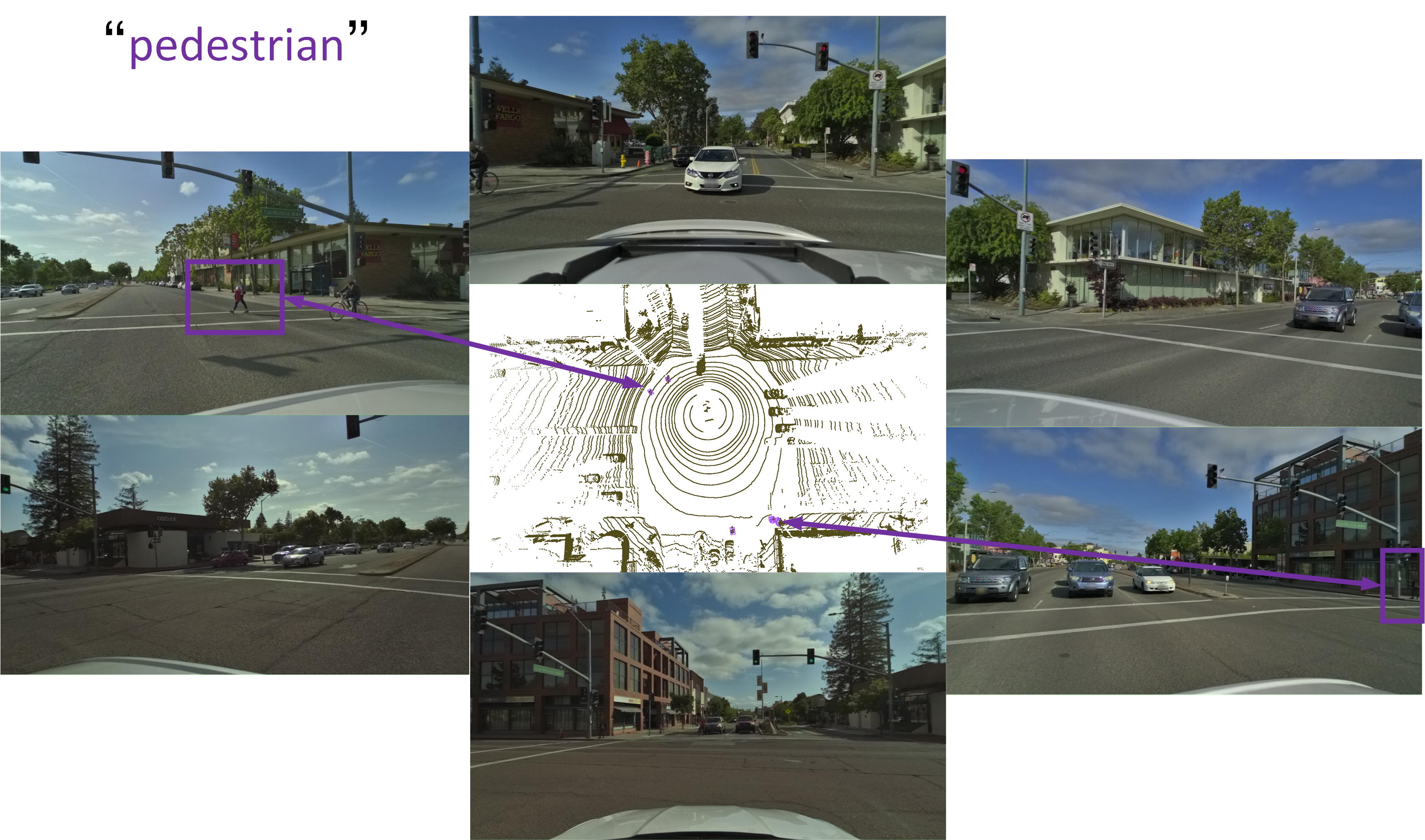}
        \label{fig:pe}
    \end{minipage}

    \caption{Lyft's zero-shot results. On the Lyft dataset, which has significant differences in data distribution, our model can locate and recognize objects specified by textual inputs. Due to space limitations, additional visualization results are provided in the appendix. Note that our model takes only point cloud and text as input, with images used solely to display the detection results.}
    \label{fig:lyft}
\end{figure*}

\paragraph{Zero-Shot Performance}
To evaluate zero-shot performance, Open3DWorld was run on the Lyft Level 5 dataset. Results \ref{table:lyft} indicated that the fusion model, leveraging BEV features and text features, can align theoretically to achieve classification.As we all know, in the field of point cloud object detection, when we use a model trained on one dataset to test on another dataset, the model will crash.Our method does not require training and can achieve preliminary results on Lyft Level 5 dataset, which shows that after the text features and BEV features are aligned, the generalization is much better than the traditional detection model.

\paragraph{Ablation Experiments}

To evaluate the scalability and versatility of our framework, we conducted ablation experiments focusing on both the text backbone and the point cloud backbone. We compared the CLIP Text Encoder and BERT-base for the text backbone, with both models kept frozen during the tests. For the point cloud backbone, we evaluated PointPillars, SSN \cite{zhu2020ssn}, CenterFormer \cite{zhou2022centerformer}, and our proposed OpenSECOND. These comparisons included the number of parameters, inference time, and key performance metrics. The results, shown in Table \ref{table:nus}, highlight the trade-off between mean average precision (mAP) and inference speed. The "Params" column refers to the total number of parameters across the entire model, including the backbone, fusion module, and heads.

Figure \ref{fig:nus} presents examples from the NuScenes-T dataset. We display 3D boxes related to the input texts on images from six surround cameras, with the corresponding BEV boxes shown on the right. The text color matches the color of the related boxes.

Figure \ref{fig:lyft} shows the zero-shot inference results on the Lyft dataset. The inputs are text and point cloud data. We use different colors to indicate different text types, showing the point cloud and detection results in the middle. The box colors match the corresponding text colors. For comparison, we include six images taken by the surround cameras.

\section{Discussion}

Key insights gained from the experiments underscored both the strengths and weaknesses of the proposed fusion model. The integration of 3D point clouds and textual data significantly enhanced detection accuracy; however, the current collection of open vocabulary categories remains limited in both quantity and quality. In contrast to the tens of thousands of categories available in the field of image-based open vocabulary detection, our dataset requires further expansion. Additionally, there is an imbalance in the proportion of different categories, which negatively impacts the model's performance, particularly on long-tail categories.

To address these challenges, future work will focus on automatically collecting open vocabulary data annotations or adopting unsupervised methods to train the model, enabling it to manage a broader range of categories in real-world scenarios. Moreover, future research will explore the integration of additional data modalities and the development of more advanced fusion techniques to further enhance object detection capabilities in autonomous driving contexts.

\section{Conclusion}

In this work, we introduced a novel approach to 3D open vocabulary detection in autonomous driving, leveraging the fusion of LIDAR and textual data. Our method enables autonomous vehicles to adapt to new textual inputs without the need for extensive retraining, thereby enhancing their ability to operate in diverse and dynamic environments. Through the integration of 3D point clouds with textual data, our approach improves object localization and identification directly from textual queries. The effectiveness of our framework has been demonstrated through extensive experiments on the NuScenes-T dataset and validated on the Lyft Level 5 dataset, showcasing its robustness and versatility in real-world scenarios. By advancing the integration of multimodal data, this work contributes to the development of safer, more reliable, and efficient autonomous driving systems, paving the way for more adaptable and intelligent perception models in the future.

\end{document}